\documentclass[10pt,twocolumn,letterpaper]{article}

\usepackage{iccv}
\usepackage{times}
\usepackage{epsfig}
\usepackage{graphicx}
\usepackage{amsmath}
\usepackage{amssymb}
\usepackage{setspace}

\usepackage[pagebackref=true,breaklinks=true,letterpaper=true,colorlinks,bookmarks=false]{hyperref}

 \iccvfinalcopy 

\def\iccvPaperID{****} 

\begin{document}

\title{Gradient Information Guided Deraining with A Novel Network and Adversarial Training}

\author{Yinglong Wang{$^{\dag\ast}$}, Haokui Zhang{$^{\S\ast}$}, Yu Liu{$^{\ddag}$}, Qinfeng Shi{$^{\ddag}$}, Bing Zeng{$^{\dag}$}\\
\thanks{$\ast$ means equal contribution.}
$^{\dag}$School of Information and Communication Engineering,\\ University of Electronic Science and Technology of China\\
$^{\ddag}$The University of Adelaide\\
$^{\S}$Northwestern Polytechnical University \\
}

\maketitle

\begin{abstract}

   In recent years, deep learning based methods have made significant progress in rain-removing. However, the existing methods usually do not have good generalization ability, which leads to the fact that almost all of existing methods have a satisfied performance on removing a specific type of rain streaks, but may have a relatively poor performance on other types of rain streaks. In this paper, aiming at removing multiple types of rain streaks from single images, we propose a novel deraining framework (GRASPP-GAN), which has better generalization capacity. Specifically, a modified ResNet-18 which extracts the deep features of rainy images and a revised ASPP structure which adapts to the various shapes and sizes of rain streaks are composed together to form the backbone of our deraining network. Taking the more prominent characteristics of rain streaks in the gradient domain into consideration, a gradient loss is introduced to help to supervise our deraining training process, for which, a Sobel convolution layer is built to extract the gradient information flexibly. To further boost the performance, an adversarial learning scheme is employed for the first time to train the proposed network. Extensive experiments on both real-world and synthetic datasets demonstrate that our method outperforms the state-of-the-art deraining methods quantitatively and qualitatively. In addition, without any modifications, our proposed framework also achieves good visual performance on dehazing.

\end{abstract}

\section{Introduction}

In real-world scenarios, many practical computer vision applications include self-driving cars \cite{Stock_2018_online}, traffic surveillance \cite{Bahnsen_2018_TITS} are heavily affected by the weather conditions, such as haze, snow and rain. Compared to the static haze condition \cite{Wang_W_2017_TMM}, the dynamic and very changeable rain streaks in shapes, sizes and intensities will cause more serious degradation on images/videos, so is the performance of many computer vision algorithms, especially these which are based on the small image features. Therefore, removing rain from images/videos is a significant problem in computer vision community and attracts attentions of more and more people.

\begin{figure}
\begin{center}
\begin{minipage}{0.32\linewidth}
\centering{\includegraphics[width=1\linewidth]{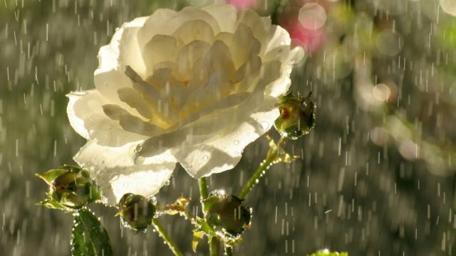}}
\centerline{Input}
\end{minipage}
\hfill
\begin{minipage}{0.32\linewidth}
\centering{\includegraphics[width=1\linewidth]{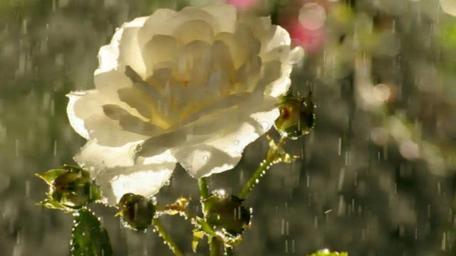}}
\centerline{DDN \cite{Fu_2017_CVPR}}
\end{minipage}
\hfill
\begin{minipage}{0.32\linewidth}
\centering{\includegraphics[width=1\linewidth]{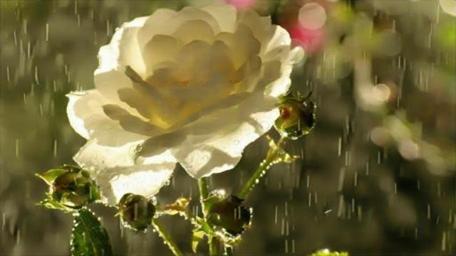}}
\centerline{DID-MDN \cite{Zhang_2018_CVPR}}
\end{minipage}
\vfill
\begin{minipage}{0.32\linewidth}
\centering{\includegraphics[width=1\linewidth]{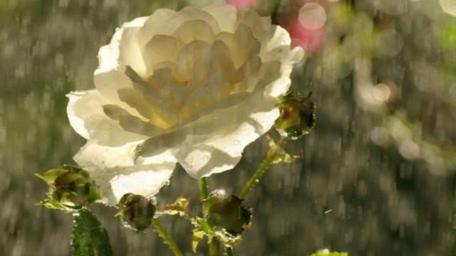}}
\centerline{RESCAN \cite{Li_2018_ECCV}}
\end{minipage}
\hfill
\begin{minipage}{0.32\linewidth}
\centering{\includegraphics[width=1\linewidth]{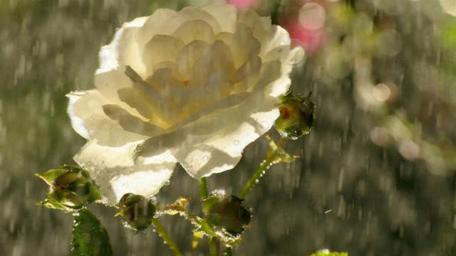}}
\centerline{JORDER \cite{Yang_2017_CVPR}}
\end{minipage}
\hfill
\begin{minipage}{0.32\linewidth}
\vskip -1.5pt
\centering{\includegraphics[width=1\linewidth]{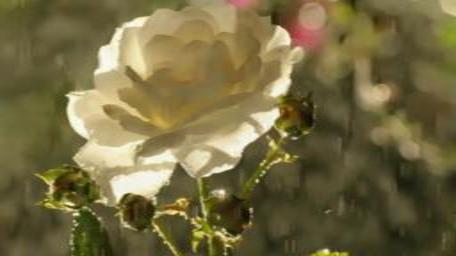}}
\centerline{GRASPP-GAN}
\end{minipage}
\end{center}
 \caption{An example of deraining results of several state-of-the-art methods and our method (GRASPP-GAN). Our method removes rain streaks more completely and obtains better visual effects.}
  \label{fig:teaser}
\end{figure}

In this paper, we focus on removing rain streaks from single color images. Early single-image deraining works usually decompose rainy images by dictionary learning technique \cite{Mairal_2010_JMLR}, then removing rain streaks by identifying the dictionary atoms which only contain rain information \cite{Fu_2011_ASSP,Chen_2014_CSVT,Huang_2014_TMM,Kang_2012_TIP,Wang_2016_ICIP,Wang_2017_TIP}. In \cite{Huang_2014_TMM}, a self-learning based dictionary learning method is proposed to further improve the performance. Dictionary learning is always time-consuming and the features which are utilized to identify rain dictionary atoms tend not to be adaptive to various rain streaks, hence the generalization of these methods are relatively low. Some other works build models for rainy images \cite{Li_2016_CVPR,Luo_2015_ICCV}, these models always mistreat some useful image details as rain streaks. 

In recent years, many deraining works based on deep learning have appeared \cite{Fu_2017_CVPR,Fu_2017_TIP,Zhang_2018_CVPR,Li_2018_ECCV,Yang_2017_CVPR,Li_2018_MM,Fan_2018_MM}. Deep learning, as a powerful tool, greatly boosts the rain-removing performance in objective assessment, visual quality and running speed.
The major limitation of these  methods is that they are usually designed for a certain type of rain streaks, thus do not obtain satisfactory performance when encountered with different types of rainy images.
It is also much more challenging  to remove the rain streaks which are wide and have blurry edges than that are bright and have clear edges.

Targeting at an algorithm which is suitable for the challenging rain streaks (are wide and have blurry edges) and also has a better generalization to different types of rain, in this paper, we propose a new network structure which is adaptive to the various shapes and sizes of rain. Besides, gradient information and adversarial training are also introduced to restore image texture details and boost deraining performance. 

\noindent
\textbf{Various shapes and sizes of rain}
Rain, as a classical dynamic weather, tends to have changeable shapes and sizes. Convolution layers which have single kernel size cannot extract the features of rain completely. In this work, we utilize Atrous Spatial Pyramid Pooling (ASPP) which is developed in DeepLab \cite{chen2018deeplab} to extract multi-scale features of rain. Compared with the common Spatial Pyramid Pooling (SPP) \cite{He_2015_arxiv}, ASPP can effectively capture multi-scale features by simply tuning the atrous rate and make the obtained multi-scale features keep the same size as the observed rainy image. Therefore, it is widely used in mid/high-level image processing tasks, such as image segmentation and depth prediction. However, for low-level deraining task, directly applying ASPP cannot work well, since some image features obtained in large receptive fields may harm the performance of such low-level vision task. Hence, some necessary changes are made on ASPP to adapt to low-level image processing, which can extract multi-scale features of rain better. The revised ASPP is combined with ResNet-18 which is utilized to extract efficient deep features of rainy images to constitute the backbone of our deraining network, named as RASPP.

\begin{figure*}[t]
\begin{center}
\includegraphics[width=7.0in]{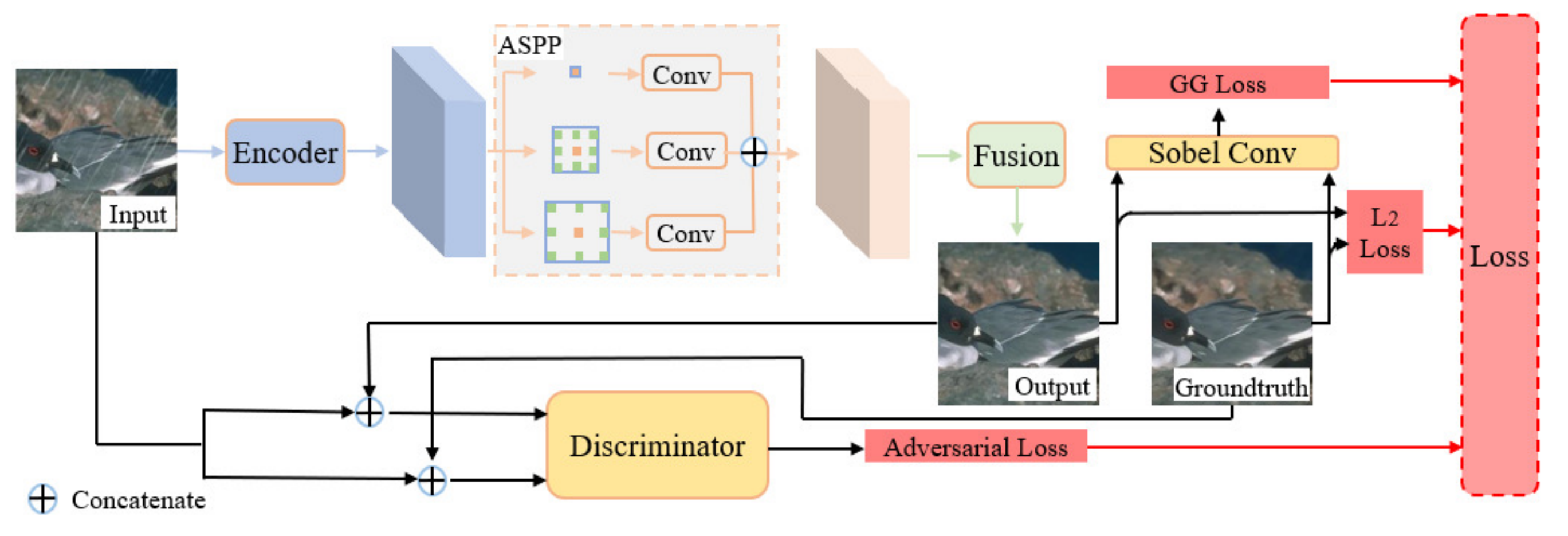}
\end{center}
   \vspace{-0.2cm}
   \caption{Illustration of our framework. The framework contains three main parts: an encoder, an ASPP module and a fusion module. In this paper, the encoder is a modified ResNet-18. The ASPP module consists of three parallel paths, each of which contains an atrous convolution layer and a pointwise convolution layer. The fusion module has three convolution layers. The loss function is made up with three parts: 2-norm loss (L2 loss), gradient information guided loss (GG loss) and adversarial loss.}
\label{fig:framework}
\end{figure*}
 
\noindent
\textbf{Texture and details}
In spatial domain of a rainy image, complex background contents can affect the deraining performance, hence some works remove rain streaks in high-frequency domain \cite{Fu_2017_CVPR,Fu_2017_TIP}.
Through extensive experiments, we find that rain information is more apparent in gradient domain, while the background information is less salient. Based on this observation, we utilize gradient information to guide the training process of RASPP, we call it GRASPP. Specifically, a Sobel gradient convolution layer is built to extract the gradient information flexibly, which is then utilized to construct a loss in gradient domain to help to optimize network parameters. The proposed loss can not only boost the deraining performance, but also make the texture details outstanding.

\noindent
\textbf{Adversarial training}
Through employing RASPP and gradient guided training, good deraining performance has been acquired. But for some rainy images, a few of fine image details may be not recovered and light traces of rain streaks may also remain in the deraining results. Hence adversarial learning is introduced to further boost the rain-removing performance and enhance visual quality (GRASPP-GAN). In particular, a discriminator is designed to determine whether there are still rain streaks remaining in the deraining results. By passing this information during the adversarial training, the parameters of generator will be further updated to produce better rain-removing results.

To sum up, we list the contributions of our work as follows:
\begin{itemize}

\item We propose a novel structure RASPP for single-image rain removal. The proposed structure first employs a modified ResNet-18 as an encoder to extract deep features of rainy images, then uses a tailored ASPP to deal with rain streaks with various sizes.

\item We design a simple but effective gradient information guided loss to help to optimize our RASPP in gradient domain which can significantly improve the deraining performance. In practice, we design a Sobel convolution layer to generate gradient maps conveniently. 

\item We design a useful adversarial loss  by  introducing  the generative adversarial learning scheme. The adversarial loss not only further improves quantitative evaluation metrics, but also generates visually pleasant results. To the best of our knowledge, this is the first time that the generative adversarial learning is employed for deraining. 

\end{itemize}
In addition, our proposed structure also has good generalization ability. Without any modifications, we apply our proposed framework to dehazing and achieve good visual effects.

\section{Related Works}
In this section, we will briefly look through the conventional as well as deep learning based methods for deraining for single rainy images.
\subsection{Conventional Methods}

The technique of using the over-complete dictionary (e.g., \cite{Mairal_2010_JMLR}) to sparsely describe the contents of an image has been widely used in rain-removing task. 
Fu \emph{et al.}\cite{Fu_2011_ASSP} firstly propose to use an over-complete dictionary to decompose a rainy image, so that some dictionary atoms only contain rain streaks while others include background information. Then some characteristics of rain (e.g., the directions of rain streaks are relatively consistent) are utilized to distinguish the rain atoms from the non-rain atoms. After that, the deraining result can be reconstructed from non-rain atoms via using orthogonal matching pursuit algorithm \cite{Mallat_1993_TSP}.

Later on, some improvements \cite{Kang_2012_TIP,Chen_2014_CSVT,Huang_2014_TMM,Luo_2015_ICCV,Wang_2016_ICIP,Wang_2017_TIP} are proposed, these methods still adopt the dictionary learning, but much more robust feature descriptors are utilized to recognize the rain atoms more accurately. 
In addition, Wang \emph{et al.} develop new descriptors (such as SVCC and PDIP) to express the common characteristics of snow and rain and design an algorithm which can remove both rain and snow \cite{Wang_2017_TIP}. However, there are two limitations about the over-complete dictionary based deraining methods. One is the time-consuming dictionary training process, and the other one is that the designed feature descriptors are not adaptive to the various rain streaks, which leads to low generalization.   

For the purpose of delivering much efficient methods for deraining, filter (e.g., \cite{He_2013_PAMI}) based methods are proposed \cite{Ding_2015_MTA}. Although fast speed has been achieved regarding to the processing time, the performance is not competitive to the over-complete dictionary based methods. Besides, Chen \emph{et al.} \cite{Chen_2013_ICCV} propose to use a low-rank appearance model to model the correlation of the rain streaks in spatial-temporal domain. Li \emph{et al.} \cite{Li_2016_CVPR} make use of the Gaussian mixture model (GMM) based layer priors to adapt to the various scales and orientations of the rain streaks. However, both of the two methods have the drawback of mis-regarding fine image details as rain streaks, which will damage the image details of the deraining results.

\subsection{Deep Learning Based Methods} 

In recent years, there are a lot of deep learning based rain-removing methods emerging, and they achieve superior performance compared to the conventional deraining methods. In summary, almost all of the existing deep learning based deraining methods learn a rain residual layer which is subtracted from the observed rainy image to acquire the deraining result \cite{Yang_2017_CVPR,Zhang_2018_CVPR}, or directly estimating the background from the rainy image \cite{Pan_2018_arxiv_Learning}.

In particular, Fu \emph{et al.} \cite{Fu_2017_TIP} propose a DerainNet which is conducting image detail extraction in the high-frequency domain rather than the image domain, which can deliver a good visual deraining effects. At the same time, a deep detail network \cite{Fu_2017_CVPR} which is inspired by the residual network (ResNet) \cite{He_2015_CVPR} is proposed. Through 
training on the high-pass domain, the mapping range between the input and output is reduced to ease the training process.
Albeit the network which is trained on the high-frequency part could ease the training process to some extend, some bright rain streaks may still remain in the deraining results.

To handle bright rain streaks better, a binary map is used by Yang \emph{et al.} to locate rain streaks. They also create a new model to represent accumulated rain streaks (these rain streaks appear as mist or haze instead) and their various shapes and directions \cite{Yang_2017_CVPR}. This method is very good for removing bright rain streaks, but often fails for removing blur rain streaks.

Zhang \emph{et al.} \cite{Zhang_2018_CVPR} propose to employ a network to estimate the rain density, and a multi-stream densely connected DID-MDN structure is then utilized to better represent the rain streaks with different sizes and shapes. Although it can handle more types of rainy streaks, blur artifacts are accompanied for some images which have fine details.

In \cite{Li_2017_arxiv}, a multi-stage network which is composed of several parallel sub-networks is delivered. Every parallel sub-network is used to model the rain streaks with certain size. Li \emph{et al.} \cite{Li_2018_ECCV} use a recurrent neural network (RNN) to remove the rain streaks stage-wisely. However, this method is not capable of dealing with the rain streaks with blurry edges.

By skip-connections, a non-locally enhanced encoder-decoder network framework is utilized to capture long-range spatial dependencies and pooling indices guided decoding is used to learn increasingly abstract feature to preserve the image details \cite{Li_2018_MM}. Different from all state-of-the-art deep learning based deraining works, we utilize novel ASPP structure to extract the multi-scale features of various rain streaks. Besides, gradient information and adversarial training are further added to boot our deraining performance.
 
\subsection{Generative Adversarial Networks}
The generative adversarial network (GAN) has been a popular research topic since it is proposed by Goodfellow \emph{et al.} in~\cite{goodfellow2014generative}. The basic idea of GAN is the training of two adversarial networks, a generator and a discriminator. During the process of adversarial training, both generator and discriminator become more robust. GANs have been widely used in various applications, such as image-to-image translation~\cite{isola2017image} and synthetic data generation~\cite{mahmood2018unsupervised}. 

The GANs have shown their great capacity and potential for computer vision tasks, however, to our knowledge, GANs have not been used in deraining. In this paper, it is employed to improve the performance and enforce the visual effect in deraining for the first time.

\section{Methodology}
To remove rain streaks in a single image, we propose a novel framework which combines a modified ResNet-18 and a revised ASPP module to form its backbone (RASPP). Besides, gradient information guided loss and a discriminator which determines whether the deraining result still contains rain streaks or not are utilized during an adversarial training (GRASPP-GAN).

In this section, we elaborate on our GRASPP-GAN. First, we introduce our whole deraining workflow; next the structure of RASPP is shown in detail, especially, the ASPP is revised to suit to the low-level deraining task; then we present our gradient information guided loss function; finally we carefully explain our adversarial learning scheme.

\subsection{GRASPP-GAN}

Our deraining framework contains three main parts: a feature encoder, an ASPP module and a multi-scale feature fusion module, as illustrated in Figure~\ref{fig:framework}. Besides three losses are utilized in our adversarial training: 2-norm loss (${L}_{2}$ loss), gradient information guided loss (GG loss) and adversarial loss. During training, the three losses are combined with different weights to update our network parameters.

Specifically, the encoder first takes a single RGB image $I$ as input and outputs deep feature map ${f}_{1}$. Then, the ASPP module employs atrous convolutions to extract multi-scale features ${f}_{2}$ from ${f}_{1}$. Finally, the fusion module acting as a decoder fuses the multi-scale feature ${f}_{2}$ and generates deraining result $d$.

During training, we adopt the following loss function:
\begin{equation}\label{eq:loss}
Loss={L}_{2}+\alpha{L}_{g}+\beta{L}_{gan},
\end{equation}
where $\alpha$ and $\beta$ are weighting coefficients. ${L}_{2}={||d-g||}_{2}$ to show the difference between the deraining result $d$ and corresponding groundtruth $g$. ${L}_{g}$ and ${L}_{gan}$ are the gradient information guided loss and adversarial loss, respectively. Their definitions are given in sections 2.3 and 2.4, respectively. 

\subsection{RASPP}

Our proposed RASPP consists of the following three parts. 

{\bf Encoder.}  Deraining problem is different from other pixel-wise tasks, such as depth estimation \cite{eigen2014depth, Fu_2017_CVPR} and segmentation \cite{chen2014semantic, chen2018deeplab, chen2017rethinking, chen2018encoder}. It does not need a very deep network. Taking the ResNet-18 \cite{He_2015_CVPR} as basic model, we set the stride of its down-sampling convolution layers to $1$ to keep the spatial feature resolution unchanged. The commonly-used zero padding method in convolution is replaced with reflection padding to get a better boundary processing result.  Rectified linear units (ReLUs) are all replaced by  leaky rectified linear units (LeakyReLU) to mitigate against dead neurons ~\cite{Bing2015Empirical}.

{\bf ASPP Module.} Inspired by the good performance of Deeplab V3+ \cite{chen2018encoder} and DORN \cite{fu2018deep}, we firstly employ atrous spatial pyramid pooling (ASPP) module to extract multi-scale features of rainy images. Original ASPP can extract features in a large receptive field (large atrous rate) which may harm the performance of low-level computer vision algorithms. Hence, we make some modifications on ASPP to adapt to our deraining task. Our ASPP module only consists of three parallel paths, each of which includes an atrous convolution and a pointwise convolution. The atrous rates are 1, 2 and 4, respectively. Compared with ASPP module used in \cite{chen2018encoder, fu2018deep}, both the number of parallel layers and strous rates are reduced in our ASPP. In addition, all the atrous convolutions are symmetrically padded to reduce the influence of $0$ on our deraining results. 

{\bf Fusion Module.} Acting as a decoder, the fusion module contains three convolution layers with $3\times3$-sized kernels, the stride is $1$ and the convolutions are also symmetrically padded. Each of the first two convolution layers is followed by a batch normalization layer and a ReLU activate function. 

\subsection{Gradient Information Guided Loss}

\begin{figure}[h]
\begin{center}
\includegraphics[width=3.5in]{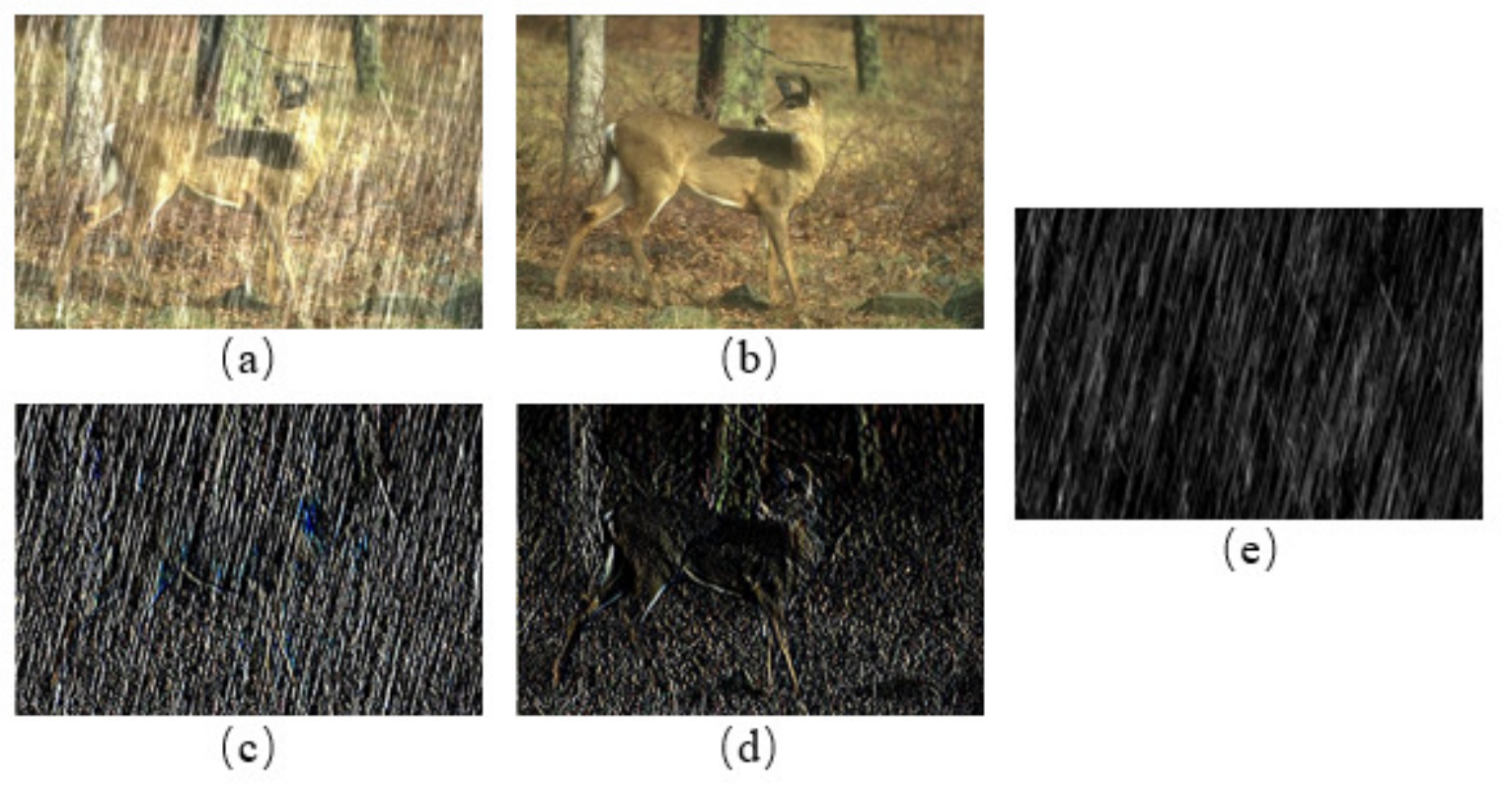}
\end{center}
   \vspace{-0.2cm}
   \caption{Comparison between spatial domain and gradient domain. (a) and (b) are the RGB images with and without rain respectively. (e) and (d) are their corresponding gradient maps along horizontal direction ($x$ axis). (e) is the rain map.}
\label{fig:gradient}
\end{figure}

In real life, looking at the same thing from another angle may make the thing easier to be solved. So far, nearly all deep learning based deraining networks are optimized by reducing the difference between the deraining result and the groundtruth in spatial domain, which can restore majority of image contents, but always neglects some fine image textures. While, these image textures tend to be more apparent in gradient domain.

Inspired by this observation, we try to solve deraining problem from two different aspects: 1) Following previous works, we optimize our network to minimize the difference between the deraining result and the corresponding groundtruth in spatial domain; 2) Taking the gradient information into consideration, we also optimize our network to minimize the differences between the deraining result and groundtruth in gradient domain. 

In fact, spatial and gradient domains are complementary. An example is shown in Figure~\ref{fig:gradient}. Figure~\ref{fig:gradient} (a) and Figure~\ref{fig:gradient} (b) are the rainy image and corresponding groundtruth, (c) and (d) are their gradient maps along horizontal direction ($x$ axis). Figure~\ref{fig:gradient} (e) is the rain layer. By comparing Figure~\ref{fig:gradient} (a)(b) with their gradient (c) and (d), we can find that spatial and gradient domains cover different information. Spatial domain mainly contains rich contents information. However, gradient domain focuses on edges and textures. In addition, for deraining task, gradient domain has several advantages that spatial domain does not have. From Figure~\ref{fig:gradient} (c), we can see more prominent structural and texture features of rain  which are difficult to capture from spatial domain.  Furthermore, the textures of rain in (c) are more close to the textures in the rain layer (e). 

Therefore, we design a gradient information guided loss to optimize our network in gradient domain. The objective function of gradient information guided loss is expressed as:

\begin{equation}
{L}_{g}= \frac{1}{n}\sum_{i=1}^{n}\left[{||{({\triangledown}_{x}({d}_{i})- {\triangledown}_{x}({g}_{i})||}_{2}+{||({\triangledown}_{y}({d}_{i})- {\triangledown}_{y}({g}_{i})||}_{2}}\right],
\end{equation}
where ${\triangledown}_{x}$ and ${\triangledown}_{y}$ represent the spatial derivative along the horizontal and vertical directions respectively. $n$ is the number of training samples. In our work, we calculate the gradient map with designed Sobel convolution layer, whose kernels are Sobel operators \cite{Kanopoulos2002Design}. By integrating gradient information guided loss to the training of our backbone RASPP, we obtain the deraining structure GRASPP.

\subsection{Adversarial Learning Scheme}

\begin{figure}[h]
\begin{center}
\includegraphics[width=2.2in]{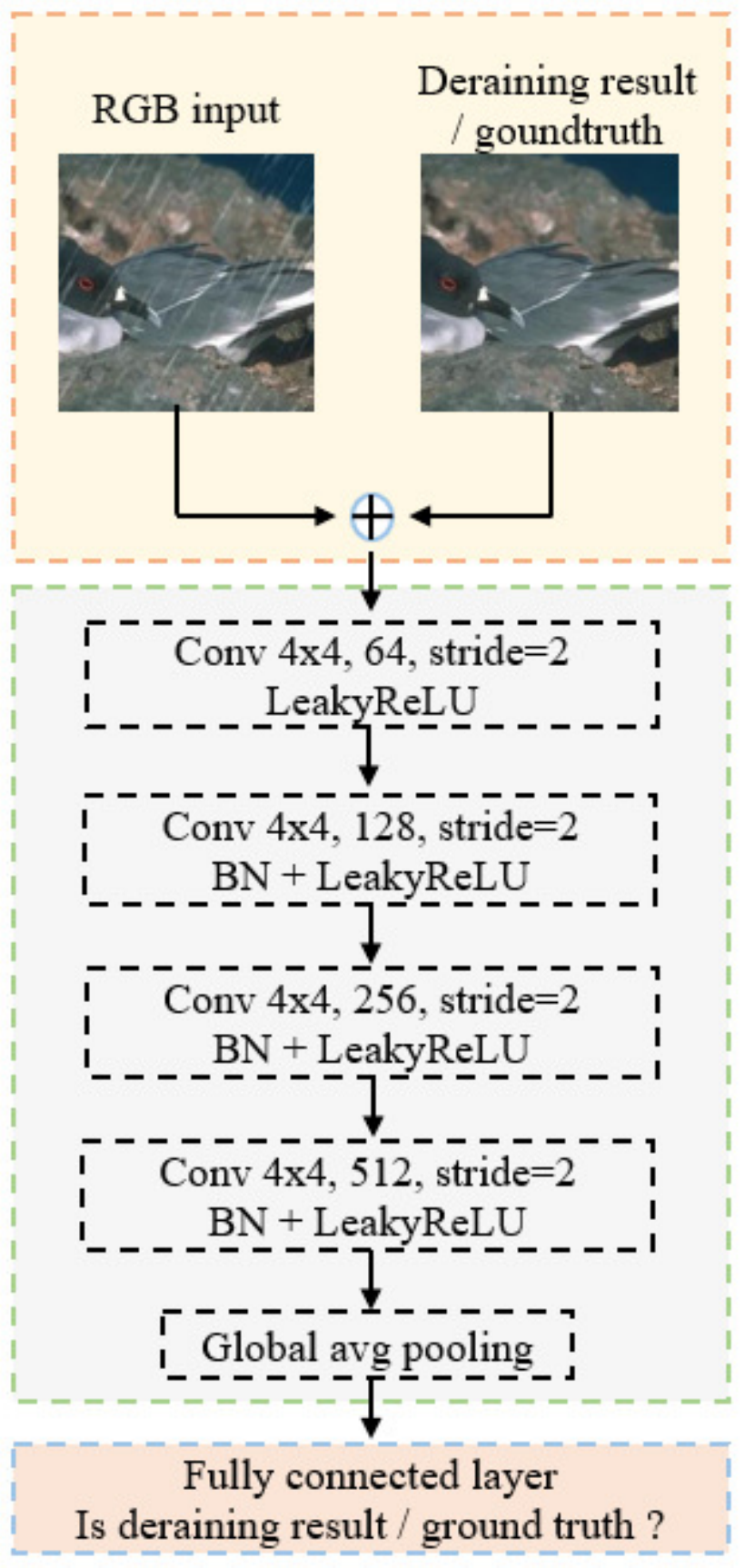}
\end{center}
   \vspace{-0.2cm}
   \caption{Structure of our discriminator. It contains 4 convolution blocks, a global average pooling layer and a fully connected layer. We concatenate the deraining result or goundtruth with the corresponding RGB input then feed the concatenated data to the discriminator. The discriminator outputs a binary label which indicates the input source.}
\label{fig:discriminator}
\end{figure}

To further improve the performance, we introduce an adversarial learning scheme which helps to train our proposed GRASPP. Specifically, after our fusion module produces the deraining results, we concatenate the results with corresponding input rainy images and feed the concatenated data to a 2D CNN discriminator to generate a score. This score denotes the probability of the concatenated data coming from our deraining network rather than the groundtruth. We train the discriminator by maximizing the probability of assigning the correct label to both the deraining results and groundtruthes. Our deraining model acts as the generator. During the generative adversarial training, we train the discriminator and generator simultaneously. The objective of our generative adversarial learning is expressed as follows:
\begin{equation}
\begin{aligned}
& \min_{G}\max_{D}V(G,D)=\\
& \mathbb{E}_{\mathbf{z}\in\zeta}[\log(D(\mathbf{z}))]+\mathbb{E}_{\mathbf{x}\in\chi}[\log(1-D(G(\mathbf{x})))],
\end{aligned}
\label{formula:temporal_loss}
\end{equation}
where $\mathbf{x}=[x_1, ..., x_n]$ are the input RGB rainy images and $\mathbf{z}=[g_1, ..., g_n]$ are the ground-truth RGB images. $\chi$ and $\zeta$ are the distributions of input images and ground-truth images, respectively. $D$ and $G$ represent discriminator and generator. Since the discriminator is a binary classifier, we use a cross entropy loss as the object function during training. 

The detailed structure of our discriminator is illustrated in Figure \ref{fig:discriminator}. It has 4 convolution blocks, a global average pooling layer and a fully-connected layer. The first convolution block consists of two layers, a convolution layer with $4\times4$ kernel and a LeakyReLU activate function. Each of the last three convolution block contains a convolution layer, followed by a batch normalization layer and a LeakyReLU layer. All convoluiton layers have a stride of 2. The formula of LeakyReLU is
\begin{equation}
LeakyReLU(x)=max(0, x) + negative_slope \ast min(0,x),
\label{formula:LeakyReLU}
\end{equation}
where $negative_slope$ is set to 0.2 in our model. In practice, as illustrated in Figure \ref{fig:discriminator}, we concatenate the deraining result or goundtruth with the corresponding rainy image, then feed them to the discriminator. During adversarial training, the adversarial loss for generator is calculated via the formula:
\begin{equation}
{L}_{gan}=\mathbb{E}_{\mathbf{x}\in\chi}\left[\log(D(G(\mathbf{x}))\right].
\label{formula:l_gan}
\end{equation}

\section{Experiments}

To evaluate the performance of our method, PSNR and SSIM \cite{Wang_2004_TIP} are utilized as our objective evaluation metrics. For qualitative assessments, we show the deraining results of some real-world and synthetic rainy images visually. As the authors of \cite{Li_2018_MM} and \cite{Fan_2018_MM} are not convenient to release their codes currently, another four state-of-the-art deraining works \cite{Fu_2017_CVPR,Zhang_2018_CVPR,Li_2018_ECCV,Yang_2017_CVPR} are selected to make comparisons.

\subsection{Implementation Details}

In the training process, we crop image patches of $128 \times 128$ from each pair of training samples. All the training image
patches are cropped randomly from original training sample pairs. We adopt Adam \cite{Kingma_2015_ICLR} to train our network and begin with the learning rate equal to $0.001$ and $0.1$ for generator and discriminator respectively, then decrease them after each epoch by multiplying
$0.1$ when the loss stops improving. 
During the training process, the generator is first trained by two epoches, then discriminator is added in the training process.
Our network is trained
on an NVIDIA 1080Ti GPU based on PyTorch. The batch sizes is 4.
The values of $\alpha$ and $\beta$ in our loss function \eqref{eq:loss} are $1$ and $0.001$ respectively.

\begin{figure*}[!t]
\begin{center}
\begin{minipage}{0.135\linewidth}
\centering{\includegraphics[width=1\linewidth]{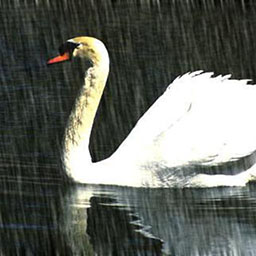}}
\end{minipage}
\hfill
\begin{minipage}{0.135\linewidth}
\centering{\includegraphics[width=1\linewidth]{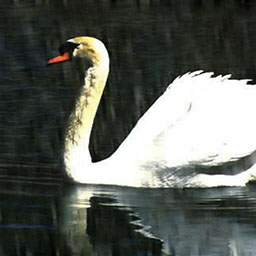}}
\end{minipage}
\hfill
\begin{minipage}{0.135\linewidth}
\centering{\includegraphics[width=1\linewidth]{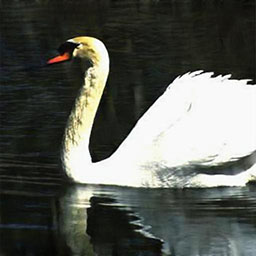}}
\end{minipage}
\hfill
\begin{minipage}{0.135\linewidth}
\centering{\includegraphics[width=1\linewidth]{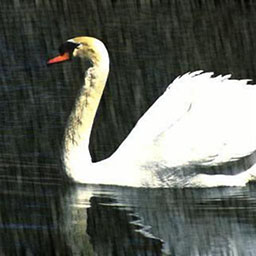}}
\end{minipage}
\hfill
\begin{minipage}{0.135\linewidth}
\centering{\includegraphics[width=1\linewidth]{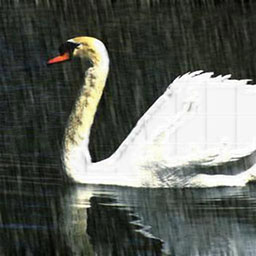}}
\end{minipage}
\hfill
\begin{minipage}{0.135\linewidth}
\centering{\includegraphics[width=1\linewidth]{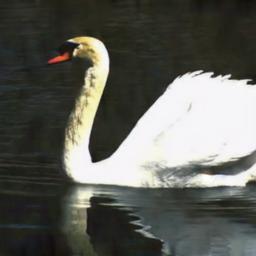}}
\end{minipage}
\hfill
\begin{minipage}{0.135\linewidth}
\centering{\includegraphics[width=1\linewidth]{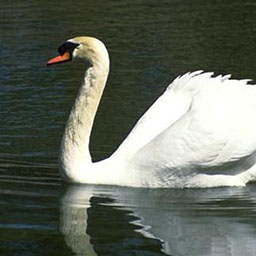}}
\end{minipage}
\vfill
\begin{minipage}{0.135\linewidth}
\centering{\includegraphics[width=1\linewidth]{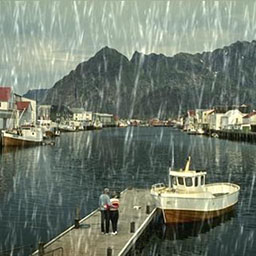}}
\end{minipage}
\hfill
\begin{minipage}{0.135\linewidth}
\centering{\includegraphics[width=1\linewidth]{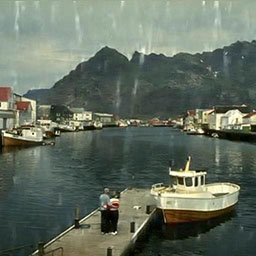}}
\end{minipage}
\hfill
\begin{minipage}{0.135\linewidth}
\centering{\includegraphics[width=1\linewidth]{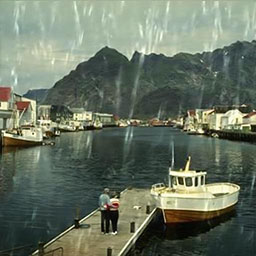}}
\end{minipage}
\hfill
\begin{minipage}{0.135\linewidth}
\centering{\includegraphics[width=1\linewidth]{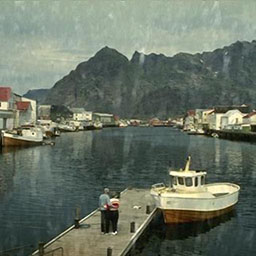}}
\end{minipage}
\hfill
\begin{minipage}{0.135\linewidth}
\centering{\includegraphics[width=1\linewidth]{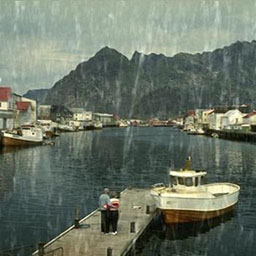}}
\end{minipage}
\hfill
\begin{minipage}{0.135\linewidth}
\centering{\includegraphics[width=1\linewidth]{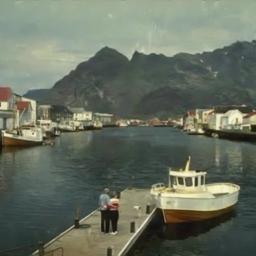}}
\end{minipage}
\hfill
\begin{minipage}{0.135\linewidth}
\centering{\includegraphics[width=1\linewidth]{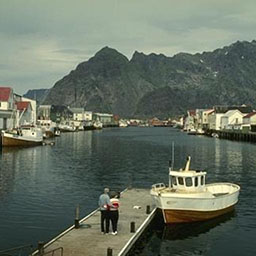}}
\end{minipage}
\vfill
\begin{minipage}{0.135\linewidth}
\centering{\includegraphics[width=1\linewidth]{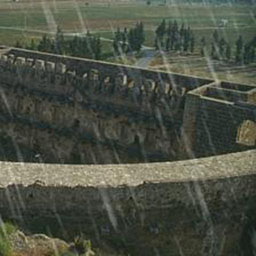}}
\centerline{Input}
\end{minipage}
\hfill
\begin{minipage}{0.135\linewidth}
\centering{\includegraphics[width=1\linewidth]{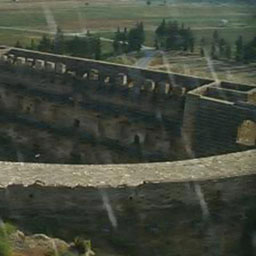}}
\centerline{DDN \cite{Fu_2017_CVPR}}
\end{minipage}
\hfill
\begin{minipage}{0.135\linewidth}
\centering{\includegraphics[width=1\linewidth]{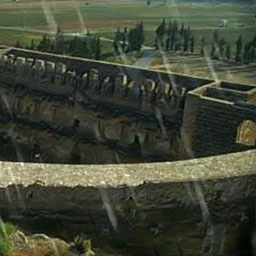}}
\centerline{DID-MDN \cite{Zhang_2018_CVPR}}
\end{minipage}
\hfill
\begin{minipage}{0.135\linewidth}
\centering{\includegraphics[width=1\linewidth]{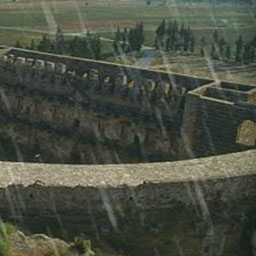}}
\centerline{RESCAN \cite{Li_2018_ECCV}}
\end{minipage}
\hfill
\begin{minipage}{0.135\linewidth}
\centering{\includegraphics[width=1\linewidth]{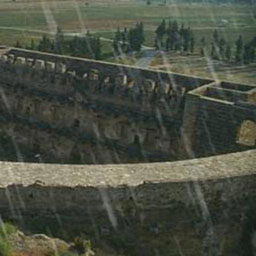}}
\centerline{JORDER \cite{Yang_2017_CVPR}}
\end{minipage}
\hfill
\begin{minipage}{0.135\linewidth}
\vskip -2pt
\centering{\includegraphics[width=1\linewidth]{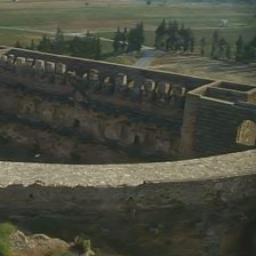}}
\centerline{GRASPP-GAN}
\end{minipage}
\hfill
\begin{minipage}{0.135\linewidth}
\vskip -2pt
\centering{\includegraphics[width=1\linewidth]{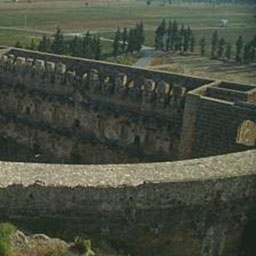}}
\centerline{GT}
\end{minipage}
\end{center}
\caption{{Visual comparisons of rain-removed results of selected state-of-the-art methods and our method on some synthetic rainy images which are from different testing datasets respectively. GT is short for groundtruth.}}
\label{fig:synthetic_comparison}
\end{figure*}

\begin{table*}[]
\centering
\caption{PSNR and SSIM comparisons of rain-removed results of selected state-of-the-art methods and our method on our three testing datasets. The selected state-of-the-art methods are trained on the training datasets which are provided in their papers.}
\label{tab:objective_compare}
\begin{tabular}{c|cc|cc|cc}
\hline
Baseline & \multicolumn{2}{c|}{Rain-I} & \multicolumn{2}{c|}{Rain-II} & \multicolumn{2}{c}{Rain-III} \\ \hline
Metric &   PSNR        &    SSIM       &     PSNR      &    SSIM       &    PSNR       &  SSIM         \\ \hline
 \cite{Fu_2017_CVPR}&   {\color{blue}29.25}        &   0.892        &    22.17       &    0.732       &    {\color{blue}29.72}       &  {\color{blue}0.887}         \\
\cite{Zhang_2018_CVPR} &    27.20       &    {\color{blue}0.898}       &    20.13       &    0.716       &    25.03       & 0.871          \\
\cite{Li_2018_ECCV} &     27.60      &    0.878       &     {\color{red}26.45}      &    {\color{red}0.845}      &    27.34       &   0.880        \\
\cite{Yang_2017_CVPR} &    28.18       &    0.853       &    23.45       &    0.749       &    29.13       & 0.881          \\ \hline
Ours &    {\color{red} 30.28}      &   {\color{red} 0.911}     &    {\color{blue}24.38}       &    {\color{blue}0.831}       &    {\color{red}31.70}       &     {\color{red}0.943}      \\ \hline
\end{tabular}
\end{table*}

\begin{table*}[]
\centering
\caption{PSNR and SSIM comparisons of rain-removed results of selected state-of-the-art methods and our method on our three testing datasets. The selected state-of-the-art methods are trained on our training dataset.}
\label{tab:objective_compare1}
\begin{tabular}{c|cc|cc|cc}
\hline
Baseline & \multicolumn{2}{c|}{Rain-I} & \multicolumn{2}{c|}{Rain-II} & \multicolumn{2}{c}{Rain-III} \\ \hline
Metric &   PSNR        &    SSIM       &     PSNR      &    SSIM       &    PSNR       &  SSIM         \\ \hline
 \cite{Fu_2017_CVPR}&   {\color{blue}27.62}        &   0.847        &    21.50       &    0.701       &    {\color{blue}30.67}       &  {\color{blue}0.912}         \\
\cite{Zhang_2018_CVPR} &    23.20       &    {\color{blue}0.897}       &    19.56       &    0.667       &    27.62       & 0.876          \\
\cite{Li_2018_ECCV} &     26.98      &    0.871       &     {\color{blue}22.91}      &    {\color{blue}0.769}      &    28.68       &   0.901          \\ \hline
Ours &    {\color{red} 30.28}      &   {\color{red} 0.911}     &    {\color{red}24.38}       &    {\color{red}0.831}       &    {\color{red}31.70}       &     {\color{red}0.943}      \\ \hline
\end{tabular}
\end{table*}

\begin{figure*}[!t]
\begin{center}
\begin{minipage}{0.16\linewidth}
\centering{\includegraphics[width=1\linewidth]{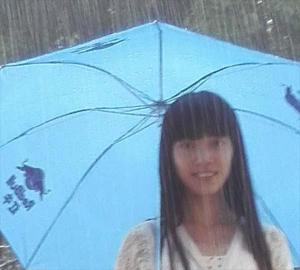}}
\end{minipage}
\hfill
\begin{minipage}{0.16\linewidth}
\centering{\includegraphics[width=1\linewidth]{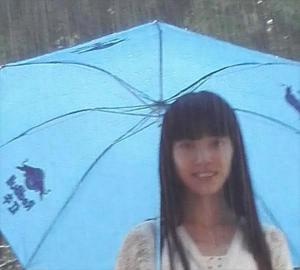}}
\end{minipage}
\hfill
\begin{minipage}{0.16\linewidth}
\centering{\includegraphics[width=1\linewidth]{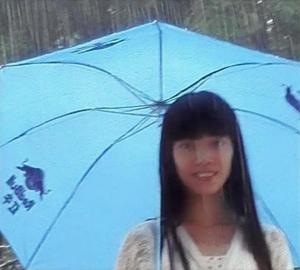}}
\end{minipage}
\hfill
\begin{minipage}{0.16\linewidth}
\centering{\includegraphics[width=1\linewidth]{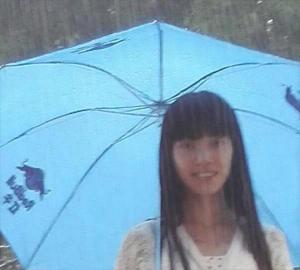}}
\end{minipage}
\hfill
\begin{minipage}{0.16\linewidth}
\centering{\includegraphics[width=1\linewidth]{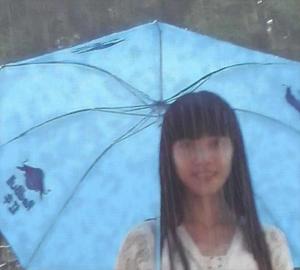}}
\end{minipage}
\hfill
\begin{minipage}{0.16\linewidth}
\centering{\includegraphics[width=1\linewidth]{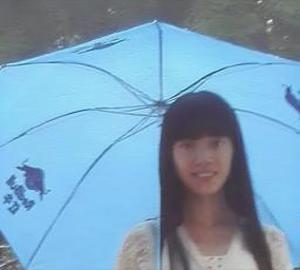}}
\end{minipage}
\vfill
\begin{minipage}{0.16\linewidth}
\centering{\includegraphics[width=1\linewidth]{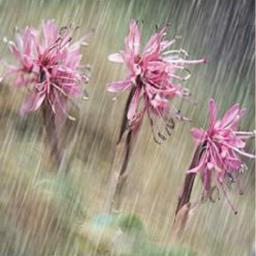}}
\end{minipage}
\hfill
\begin{minipage}{0.16\linewidth}
\centering{\includegraphics[width=1\linewidth]{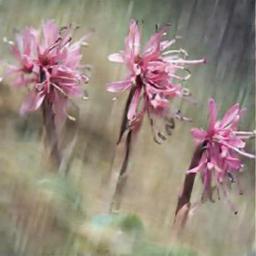}}
\end{minipage}
\hfill
\begin{minipage}{0.16\linewidth}
\centering{\includegraphics[width=1\linewidth]{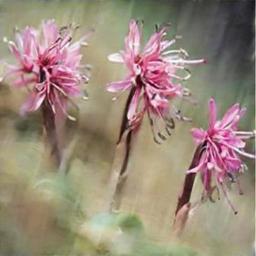}}
\end{minipage}
\hfill
\begin{minipage}{0.16\linewidth}
\centering{\includegraphics[width=1\linewidth]{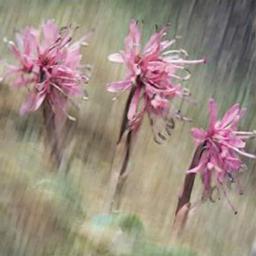}}
\end{minipage}
\hfill
\begin{minipage}{0.16\linewidth}
\centering{\includegraphics[width=1\linewidth]{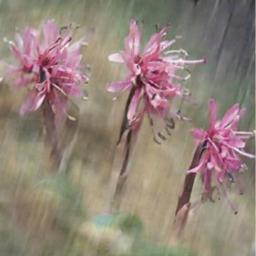}}
\end{minipage}
\hfill
\begin{minipage}{0.16\linewidth}
\centering{\includegraphics[width=1\linewidth]{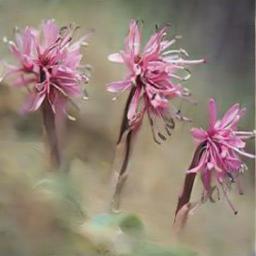}}
\end{minipage}
\vfill
\begin{minipage}{0.16\linewidth}
\centering{\includegraphics[width=1\linewidth]{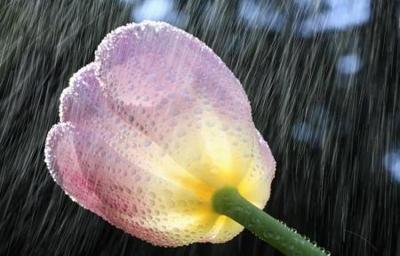}}
\centerline{Input}
\end{minipage}
\hfill
\begin{minipage}{0.16\linewidth}
\centering{\includegraphics[width=1\linewidth]{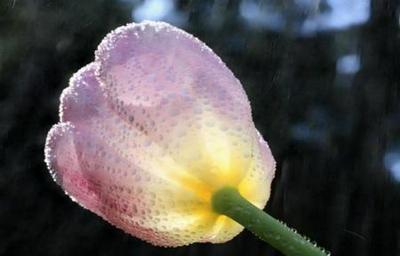}}
\centerline{DDN \cite{Fu_2017_CVPR}}
\end{minipage}
\hfill
\begin{minipage}{0.16\linewidth}
\centering{\includegraphics[width=1\linewidth]{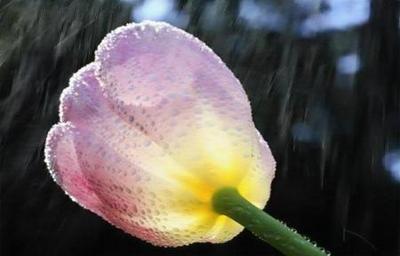}}
\centerline{DID-MDN \cite{Zhang_2018_CVPR}}
\end{minipage}
\hfill
\begin{minipage}{0.16\linewidth}
\centering{\includegraphics[width=1\linewidth]{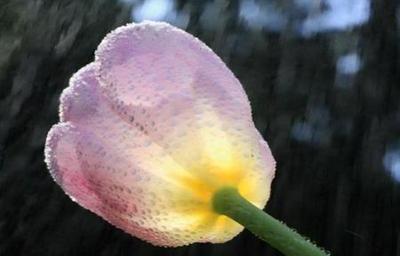}}
\centerline{RESCAN \cite{Li_2018_ECCV}}
\end{minipage}
\hfill
\begin{minipage}{0.16\linewidth}
\centering{\includegraphics[width=1\linewidth]{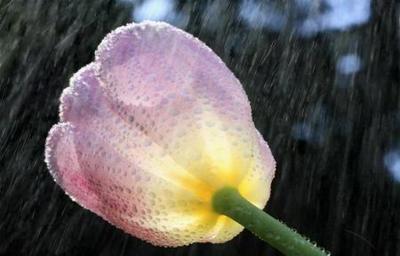}}
\centerline{JORDER \cite{Yang_2017_CVPR}}
\end{minipage}
\hfill
\begin{minipage}{0.16\linewidth}
\vskip -2pt
\centering{\includegraphics[width=1\linewidth]{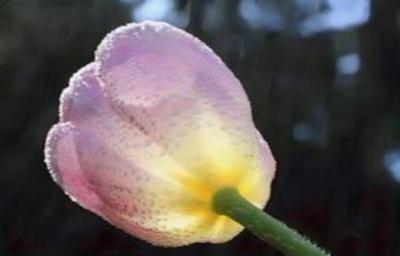}}
\centerline{GRASPP-GAN}
\end{minipage}
\end{center}
\caption{{Visual comparisons of rain-removed results of selected state-of-the-art methods and our method on some real-world rainy images.}}
\label{fig:practical_comparison}
\end{figure*}

\begin{table*}[]
\small
\newcommand{\tabincell}[2]{\begin{tabular}{@{}#1@{}}#2\end{tabular}}
\centering
\caption{Average running time on $512 \times 512$ image of different methods on our testing datasets.}
\begin{tabular}{c|c|c|c|c|c}
\hline
Methods      & DDN \cite{Fu_2017_CVPR}~(CPU) & JORDER \cite{Yang_2017_CVPR}~(GPU) & DID-MDN  \cite{Zhang_2018_CVPR}~(GPU) & RESCAN  \cite{Li_2018_ECCV}~(GPU)  & GRASPP-GAN~(GPU) \\
\hline
Time  & 0.09s & 1.40s & 0.06s & 0.50s & 0.04s      \\
\hline
\end{tabular}
\label{tab:time}
\end{table*}

\subsection{Datasets}

Majority of deraining works have their own training datasets. The datasets of \cite{Li_2018_arxiv} and \cite{Zhang_2018_CVPR} include $20800$ and $12000$ training samples respectively. In our work, we just randomly select $4000$ training pairs from these two datasets respectively to constitute our training dataset (contains $8000$ training pairs). To compare deraining  performance and generalization fairly with selected deraining works, we randomly select $100$ testing samples from the testing datasets of \cite{Fu_2017_CVPR,Li_2018_ECCV,Zhang_2018_CVPR} respectively to form our first $300$-sample testing dataset Rain-I. Yang \emph{et al.} synthesized two datasets Rain100L and Rain100H which contain $100$ samples respectively. In our work, we select Rain100H whose rain streaks are really bright as our second testing dataset Rain-II. Besides, rain streaks which are wide and have blurry edges are always difficult to be removed. Hence, another $400$-sample dataset which contains this type of rain streaks is utilized as our third dataset Rain-III.

Majority of real-world rainy images in our work are from selected state-of-the-art deraining works. Some of them are downloaded from google. Their contents are also various, including landscapes, cities, faces and so on.

\subsection{Evaluation  on Objective  Metrics}

Table \ref{tab:objective_compare} shows the PSNR and SSIM values of selected and our methods on our three datasets, and the selected state-of-the-art methods are trained on the training datasets which are provided in their own papers respectively. On Rain-I and Rain-III, our method outperforms the state-of-the-art in both PSNR and SSIM metrics. Due to the guidance of gradient, our SSIM values are remarkable and PSNR are more than $1$ db higher than the second best results of selected methods. Our objective indexes on Rain-II are second best compared with the selected methods. The best results on this dataset come from \cite{Li_2018_ECCV} by Li \emph{et al.}, but this work and the work \cite{Yang_2017_CVPR} by Yang \emph{et al.} are trained on the training dataset in \cite{Yang_2017_CVPR} which has the same rain streak type as the rain streaks in Rain-II. Though our method is not the best on Rain-II, taking the fact that our method just contains $8000$ samples compared with \cite{Zhang_2018_CVPR,Fu_2017_CVPR} which have larger training datasets, and our method is not trained on the training dataset for \cite{Yang_2017_CVPR} specially into consideration, our deraining method has better rain-removing performance, more importantly, better generalization. We will also see in the next subsection, our method obtains better deraining visual effect, especially for some challenging rainy images.

To compare more fairly, we also trained the selected state-of-the-art methods on our own training dataset, and the results are shown in Table \ref{tab:objective_compare1}. Note that the method by Yang \emph{et al.} \cite{Yang_2017_CVPR} cannot be trained on our dataset, because its training needs the guidance of location masks which are only available in their training dataset. Hence, Table \ref{tab:objective_compare1} contains the results of other three selected works and our work. We can see that when trained on our dataset, our method produces the best objective performance.

\subsection{Evaluation  on Visual Quality}

In Figure \ref{fig:synthetic_comparison}, we show some deraining results on synthetic rainy images visually. The three images come from our three testing datasets respectively. We can see that the deraining effects and generalization of selected methods are relatively low. They tend to have good results on their own testing datasets (referring to their papers), and the performance decreases when encountered with other data. The method \cite{Fu_2017_CVPR} and \cite{Zhang_2018_CVPR} are good at handling the rainy images from Rain-I dataset, in which the rain streaks are relatively light and slim. But when encountered the rain streaks under extreme case (Rain-II) or rain streaks which are wide, these two methods cannot obtain good performance. The methods \cite{Li_2018_ECCV,Yang_2017_CVPR} are specially trained on the training dataset corresponding to the extreme rain streaks (testing dataset Rain-II). Method \cite{Li_2018_ECCV} removes majority of rain streaks for the second rainy image (from Rain-II), but method \cite{Yang_2017_CVPR} does not. Both of these two methods cannot obtain good rain-removing results when rain streaks are light and dense (e.g. the first rainy image). Besides, all the selected methods are not good at handling wide rain streaks, we can see from the third synthetic rainy image, the slim rain streaks are removed but the wide rain streaks are ignored.
By comparison, our method obtains better visual quality. Majority of rain streaks are removed and the image details are also preserved well. The visual results also prove the better generalization of our method than selected methods.

The deraining results of three challenging real-world rainy images are shown in Figure \ref{fig:practical_comparison}. The first rainy image contains face which is covered by some rain streaks. Though the rain streaks in this image are light, they have blurry edges which are always challenging. The second rainy image has wide and bright rain streaks. The rain streaks in the third rainy image are relatively slim, but some rain streaks accumulate together.
Except that method \cite{Fu_2017_CVPR} produces good deraining result for the third rainy image, the selected methods cannot obtain good results for these three rainy images. By comparison, our method produces better rain-removed results and generalization. Of course, selected methods have their own advantages on some specific type of rainy images, e.g., \cite{Li_2018_ECCV} has good performance for the extreme rain streaks (Rain-II).

\begin{table*}[]
\centering
\caption{PSNR and SSIM comparisons of rain-removed results of ablation experiments on our three testing datasets.}
\label{tab:ablation_objective}
\begin{tabular}{c|cc|cc|cc}
\hline
Variants & \multicolumn{2}{c|}{RASPP} & \multicolumn{2}{c|}{GRASPP} & \multicolumn{2}{c}{GRASPP-GAN} \\ \hline
Metric &    PSNR       &    SSIM       &    PSNR       &    SSIM       &    PSNR       &        SSIM   \\ \hline
Rain-I &   28.24        &   0.886        &     {\color{blue}29.31}       &     {\color{blue}0.908}       &    {\color{red}30.28}        &    {\color{red}0.911}        \\
Rain-II &    22.58      &    0.781       &    {\color{blue}23.86}       &    {\color{blue}0.826}       &     {\color{red}24.38}       &  {\color{red}0.831}         \\
Rain-III &  29.16        &     0.920       &    {\color{blue}30.90}        &    {\color{blue}0.939}      &    {\color{red}31.70}       &  {\color{red}0.943}         \\ \hline
\end{tabular}
\end{table*}

\begin{figure*}[!t]
\begin{center}
\begin{minipage}{0.19\linewidth}
\centering{\includegraphics[width=1\linewidth]{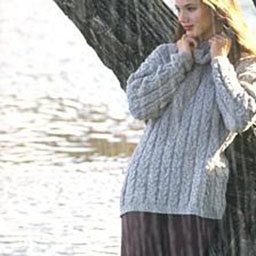}}
\end{minipage}
\hfill
\begin{minipage}{0.19\linewidth}
\centering{\includegraphics[width=1\linewidth]{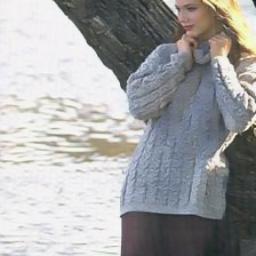}}
\end{minipage}
\hfill
\begin{minipage}{0.19\linewidth}
\centering{\includegraphics[width=1\linewidth]{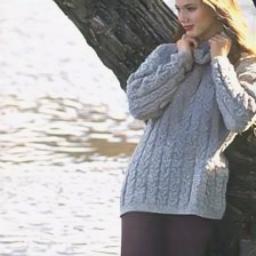}}
\end{minipage}
\hfill
\begin{minipage}{0.19\linewidth}
\centering{\includegraphics[width=1\linewidth]{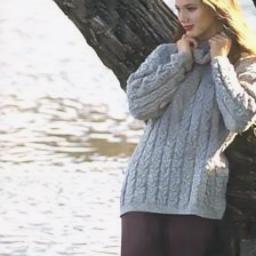}}
\end{minipage}
\hfill
\begin{minipage}{0.19\linewidth}
\centering{\includegraphics[width=1\linewidth]{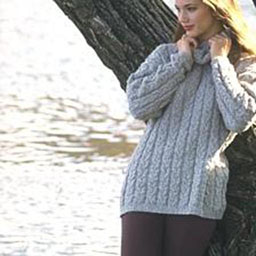}}
\end{minipage}
\vfill
\begin{minipage}{0.19\linewidth}
\centering{\includegraphics[width=1\linewidth]{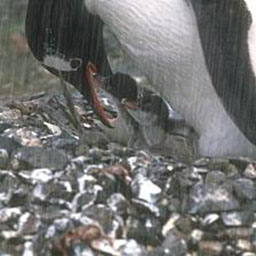}}
\centerline{Input}
\end{minipage}
\hfill
\begin{minipage}{0.19\linewidth}
\centering{\includegraphics[width=1\linewidth]{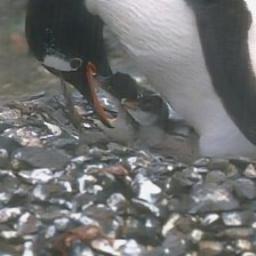}}
\centerline{RASPP}
\end{minipage}
\hfill
\begin{minipage}{0.19\linewidth}
\centering{\includegraphics[width=1\linewidth]{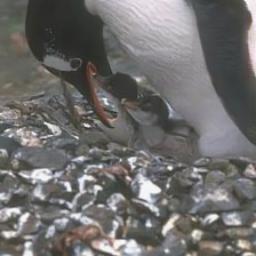}}
\centerline{GRASPP}
\end{minipage}
\hfill
\begin{minipage}{0.19\linewidth}
\centering{\includegraphics[width=1\linewidth]{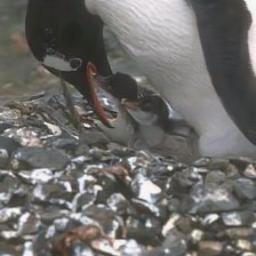}}
\centerline{GRASPP-GAN}
\end{minipage}
\hfill
\begin{minipage}{0.19\linewidth}
\centering{\includegraphics[width=1\linewidth]{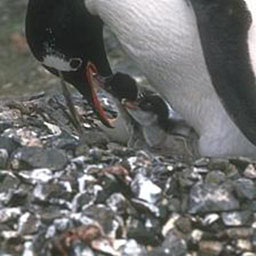}}
\centerline{GT}
\end{minipage}
\end{center}
\caption{{Visual comparisons of rain-removed results of ablation experiments on synthetic rainy images. GT is short for groundtruth.}}
\label{fig:ablation_synthetic}
\end{figure*}

\begin{figure*}[!t]
\begin{center}
\begin{minipage}{0.24\linewidth}
\centering{\includegraphics[width=1\linewidth]{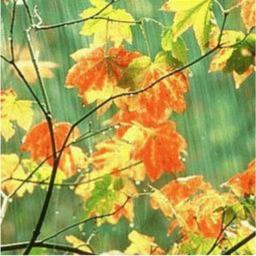}}
\end{minipage}
\hfill
\begin{minipage}{0.24\linewidth}
\centering{\includegraphics[width=1\linewidth]{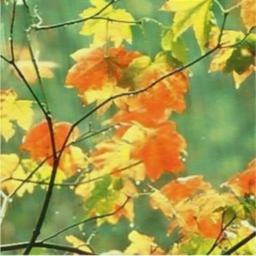}}
\end{minipage}
\hfill
\begin{minipage}{0.24\linewidth}
\centering{\includegraphics[width=1\linewidth]{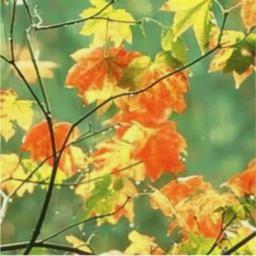}}
\end{minipage}
\hfill
\begin{minipage}{0.24\linewidth}
\centering{\includegraphics[width=1\linewidth]{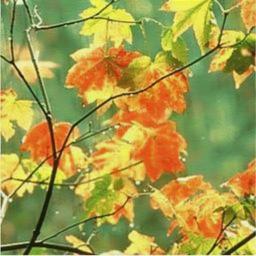}}
\end{minipage}
\vfill
\begin{minipage}{0.24\linewidth}
\centering{\includegraphics[width=1\linewidth]{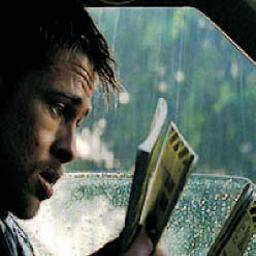}}
\centerline{Input}
\end{minipage}
\hfill
\begin{minipage}{0.24\linewidth}
\vskip -2pt
\centering{\includegraphics[width=1\linewidth]{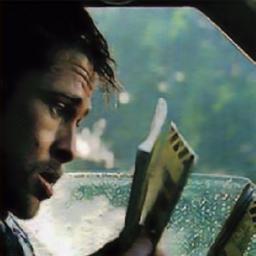}}
\centerline{RASPP}
\end{minipage}
\hfill
\begin{minipage}{0.24\linewidth}
\vskip -2pt
\centering{\includegraphics[width=1\linewidth]{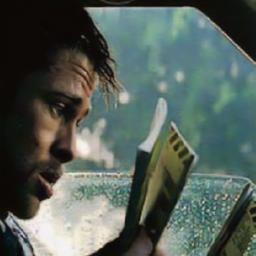}}
\centerline{GRASPP}
\end{minipage}
\hfill
\begin{minipage}{0.24\linewidth}
\vskip -2pt
\centering{\includegraphics[width=1\linewidth]{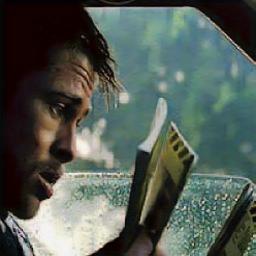}}
\centerline{GRASPP-GAN}
\end{minipage}
\end{center}
\caption{{Visual comparisons of rain-removed results of ablation experiments on real-world rainy images.}}
\label{fig:ablation_practical}
\end{figure*}

\begin{figure*}[!t]
\begin{center}
\begin{minipage}{0.24\linewidth}
\centering{\includegraphics[width=1\linewidth]{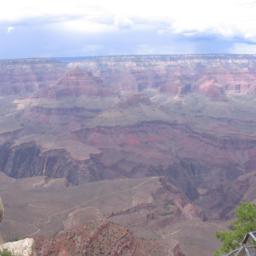}}
\end{minipage}
\hfill
\begin{minipage}{0.24\linewidth}
\centering{\includegraphics[width=1\linewidth]{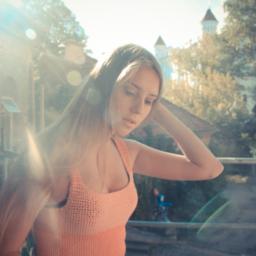}}
\end{minipage}
\hfill
\begin{minipage}{0.24\linewidth}
\centering{\includegraphics[width=1\linewidth]{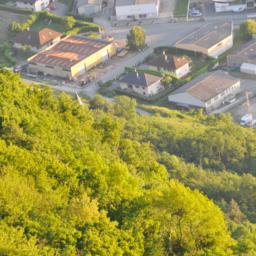}}
\end{minipage}
\hfill
\begin{minipage}{0.24\linewidth}
\centering{\includegraphics[width=1\linewidth]{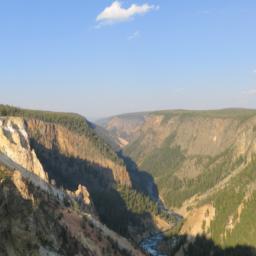}}
\end{minipage}
\vfill
\begin{minipage}{0.24\linewidth}
\centering{\includegraphics[width=1\linewidth]{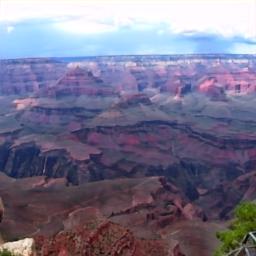}}
\end{minipage}
\hfill
\begin{minipage}{0.24\linewidth}
\centering{\includegraphics[width=1\linewidth]{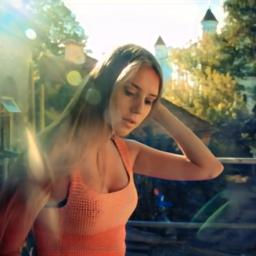}}
\end{minipage}
\hfill
\begin{minipage}{0.24\linewidth}
\centering{\includegraphics[width=1\linewidth]{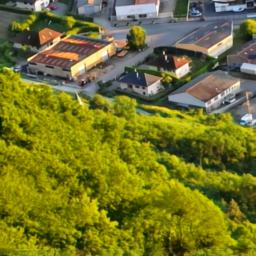}}
\end{minipage}
\hfill
\begin{minipage}{0.24\linewidth}
\centering{\includegraphics[width=1\linewidth]{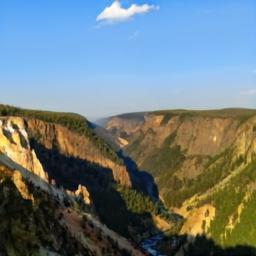}}
\end{minipage}
\end{center}
\caption{{Visual dehazing results by our network on real-world hazy images. The first line is the hazy images, and the second line is the dehazing results.}}
\label{fig:haze_practical}
\end{figure*}

\vspace{-2 mm}
\subsection{Running Time}

To illustrate the efficiency of implementing our GRASPP-GAN in practice, we show the average running time of our method on our three testing datasets in Table \ref{tab:time}, and make comparisons with selected deraining methods.
The size of all our testing rainy images is $512 \times 512$. For the work \cite{Fu_2017_CVPR}, we only obtain its CPU-version code, so we show its time running on CPU. The other methods are all tested on GPU. Because \cite{Fu_2017_CVPR} is run on CPU, it costs much more time. For other methods which are run on GPU, \cite{Yang_2017_CVPR} is the slowest one. Our method is the fastest one. 

\section{Ablation Study}

Our network utilizes the combination of ResNet-18 and ASPP as the backbone of our network, then gradient information is utilized to guide the training of our network to preserve clearer texture. At last, a discriminator is added to identify whether an image contains rain streaks or not. By adversarial training process, the performance of our method is enhanced further. Hence, we can do another two ablation experiments to prove the effectiveness of the backbone, the gradient and the discriminator in our network. The first ablation experiment is that gradient information and discriminator are both removed, only the backbone is trained on our training dataset, we call this structure RASPP. The second is that we add gradient information guidance during the training process, and we call it GRASPP. When the discriminator and adversarial training are added, the structure is our whole network GRASPP-GAN.

In Table \ref{tab:ablation_objective}, we show the PSNR and SSIM values of RASPP, GRASPP and the whole network GRASPP-GAN on our three testing datasets. Although without gradient and discriminator, from the PSNR and SSIM of oblation experiment RASPP we can see that the deraining performance of the backbone of our network is good and even better than the performance of some selected state-of-the-art works (e.g. \cite{Zhang_2018_CVPR}). This is because ResNet can effectively constrain gradient elimination during the propagation and obtain good deep features of rainy images. By ASPP structure, we can acquire multi-scale deep features which adapt to the various shapes and sizes of rain streaks well. By introducing gradient information, namely, the ablation experimant GRASPP, the PSNR and SSIM are consistently improved on our three testing datasets. During the training, we minimize not only the content difference between the deraining result and groundtruth but also minimize their gradient difference, which can preserve image details and enhance the structural similarity. Adversarial training tends to play significant role in performance improvement. By judging whether the rain-removed result still contains rain streaks or not, the discriminator feedbacks relative information to the generator to help it produce better deraining result through adversarial training. Hence, the performance of our GRASPP-GAN is enhanced further.

In Figure \ref{fig:ablation_synthetic} and \ref{fig:ablation_practical}, we visually show some results of ablation experiments on some synthetic and real-world rainy images respectively. For majority of rainy images, our GRASPP-GAN and its two ablation variants can obtain good rain-removed visual results (e.g., the second image in Figure \ref{fig:ablation_synthetic} and the two real-world rainy images in Figure \ref{fig:ablation_practical}). However, there are still some differences for some rainy images. For example, the first rainy image in Figure \ref{fig:ablation_synthetic}, we find the ablation experiment RASPP losses more image details, and the details are recovered with the introducing of gradient information and adversarial training, which shows the role of gradient and adversarial training visually. For other rainy images in Figure \ref{fig:ablation_synthetic} and \ref{fig:ablation_practical}, all these three experiments preserve good image details and the rain streaks are also removed clearly.

\section{Extension}

In this section, we simply show the potentials of our network in dehazing task. Without any changes on the structure of our network, we train it on a $5000$-sample dehazing training dataset whose sample pairs are randomly selected from the training dataset of the dehazing work \cite{Li_2019_TIP}. In Figure \ref{fig:haze_practical}, we show some dehazing results by our network on real-world hazy images. We can see that the dehazing visual quality of our network are satisfactory, and the image details are also preserved well.

\section{Conclusion}
In this paper, we proposed a gradient information guided generative adversarial network (GRASPP-GAN) for removing the rain streaks in single rainy images.
ResNet structure which can efficiently constrain gradient elimination in the propagation is utilized to extract deep features of rainy images. In order to adapt to the changeable shapes and sizes of rain streaks, a revised ASPP structure is employed to acquire multi-scale deep features which are fused to obtain rain-removed results. Rain streaks have more apparent features than the background in gradient domain. Hence, we introduced a gradient loss, for which a Sobel gradient convolution layer is built, to additionally supervise our training process and the performance is enhanced apparently. Finally, the deraining performance and visual quality are further enhanced by a discriminator which promotes the parameter optimization of the generator by determining whether the rain-removed results still contain rain streaks. Extensive experiments prove that our network outperforms the selected state-of-the-art methods in deraining performance and generalization.


\end{document}